# On the Use of Emojis to Train Emotion Classifiers

Wegdan A. Hussien,[*] Mahmoud Al-Ayyoub,[*] Yahya M. Tashtoush,[*] and Mohammed N. Al-Kabi[†]

[*] Jordan University of Science and Technology, Irbid, Jordan
Emails: wahussien13@cit.just.edu.jo, {maalshbool, yahya-t}@just.edu.jo
[†] Al-Buraimi University College, Buraimi, Sultanate of Oman
Email: mohammed@buc.edu.om

*Abstract—* **Nowadays, the automatic detection of emotions is employed by many applications in different fields like security informatics, e-learning, humor detection, targeted advertising, etc. Many of these applications focus on social media and treat this problem as a classification problem, which requires preparing training data. The typical method for annotating the training data by human experts is considered time consuming, labor intensive and sometimes prone to error. Moreover, such an approach is not easily extensible to new domains/languages since such extensions require annotating new training data. In this study, we propose a distant supervised learning approach where the training sentences are automatically annotated based on the emojis they have. Such training data would be very cheap to produce compared with the manually created training data, thus, much larger training data can be easily obtained. On the other hand, this training data would naturally have lower quality as it may contain some errors in the annotation. Nonetheless, we experimentally show that training classifiers on cheap, large and possibly erroneous data annotated using this approach leads to more accurate results compared with training the same classifiers on the more expensive, much smaller and error-free manually annotated training data. Our experiments are conducted on an in-house dataset of emotional Arabic tweets and the classifiers we consider are: Support Vector Machine (SVM), Multinomial Naive Bayes (MNB) and Random Forest (RF). In addition to experimenting with single classifiers, we also consider using an ensemble of classifiers. The results show that using an automatically annotated training data (that is only one order of magnitude larger than the manually annotated one) gives better results in almost all settings considered.**

1. INTRODUCTION

Over the last two decades, the Internet evolution has led to a transition from the static web to a dynamic web, where the users can get information and share their opinions, ideas, knowledge and even media content [1]. Consequently, social networks emerged and many social services have grown rapidly and attracted millions of people in a short time. Hence, these social networks become an important part in people's lives as they use them to communicate and share their daily events creating an amazing source of raw data that can mined for commercial as well as non-commercial purposes. Some of the lucrative ways to mine this data is to determine the users' sentiments or emotions giving rise to the fields of Sentiment Analysis (SA) and Emotion Analysis (EA).

As their names suggest, SA and EA are concerned with determining the sentiments and emotions, respectively, conveyed in a piece of text. Both are usually treated as binary or multiclass supervised learning problems with training and testing datasets. For SA, the considered sentiments are usually (positive vs negative) or (positive vs neutral vs negative). As for EA, most of the current work in the literature either follow Ekman's [2] six basic emotions (Anger, Disgust, Fear, Joy, Sadness and Surprise) or Plutchik's [3] eight basic emotions (which are Ekman's six emotions, in addition to anticipation and trust).

The field of EA is quite different from the popular field of SA. One might view some emotions as sentiments on a finer-granularity. As an example, consider the 'sadness' and 'anger' emotions. People can easily consider them as negative sentiments, but clearly, they are different emotions. EA can thus be done as an additional layer on top of the (relatively) simpler SA. Another difference is related to how many classes each problem considers and whether a text excerpt can bear more than one class or not (multi-label classification). While SA can be binary, ternary or multi-way (e.g., a 5-star ranking system), EA is much more open to interpretation. Although only six or eight emotions are usually considered basic/primary emotions, the number of different emotions (including secondary and tertiary) considered by EA can be much larger. Dealing with a larger number of classes increases the complexity of the problem. Moreover, the boundary between two emotions may not be clear. In fact, the same text excerpt might convey different emotions at the same time making it a multi-label text classification problem.[1] On the other hand, in the field of SA, the issue of conveying different sentiments (conflicting or mixed sentiments) is usually ignored.

The previous paragraph highlights the challenges faced by the EA research. One very concerning issue is that even human experts might have some difficulty determining the possibly overlapping emotions of a certain text. This means that the process preparing the training and testing dataset (which is basically the first step in this field) can be quite challenging. The problem is amplified by the fact that a very large training dataset is need to allow the classifiers to capture the nuance differences between the different emotion and their overlapping nature. Thus, instead of going through a very costly and lengthy process with human annotators, we propose to resort to automatic methods for creating the training dataset. Automatic methods for creating the training dataset might be less accurate than the manual method, however, they allow creating very large training datasets, which, intuitively, might give better overall results despite being of lowed quality. This is known as Distant Supervised (DS) learning [33].

DS learning is based on the idea of automatically determining attributes or labels of data based on possibly noisy information that are automatically extracted. Determining the emotion or sentiment of a piece of text based on the emojis used therein is an example of DS learning. Other examples include the process of synthetically generating noisy parallel corpus used in the Automatic Machine Translation (AMT) literature for under-resourced languages. Obviously, data labeled using automatic methods cannot be guaranteed to be of superior quality compared with data labeled by humans. However, the very low cost of automatically labeling the data allows us to create datasets with massive sizes at very low cost. The intuition here is that the difference in size between the automatically labeled data and the manually labeled one will compensate for the lower quality in the labeling process allowing the former to be more useful for the purposes of creating or expanding training datasets [4, 5, 9, 33-42].

In this work, we aim at showing that training a classifier to detect emotions on a large number of automatically annotated tweets (based on emojis) is actually better than training it on a smaller number of manually annotated tweets. To do so, we collect emotion-bearing Arabic tweets focusing on four basic emotions: anger, disgust, joy and sadness. We create two training sets: one with the tweets containing emojis (to be annotated automatically) and one with randomly selected tweets (to be annotated manually). We also extract a third set to be used for testing. The testing set is manually annotated. We train the same classifiers on the two training sets and test them on the same testing set. The results show that the classifiers trained on automatically trained dataset are better than the ones trained on the manually annotated dataset.

In a preliminary version of this work [45], we demonstrated a proof of concept on two classifiers: Support Vector Machine (SVM) and Multinomial Naive Bayes (MNB). In this work, we consider a third classifier, which is Random Forest (RF), and improve the classification accuracy of both approaches by exploring the idea of combining classifiers using different techniques. We also perform more extensive experiments on issues like effect of preprocessing, using only single-emoji tweets, etc.

Our results and findings are significant as they might change the way researchers approach problems like EA of social media content. They will allow them to use much larger data sizes and venture into new domains making their systems more adaptable, practical and trustworthy. They might even pave the way to building classifiers for poorly resourced languages/domains. In fact, this is one of the reasons why we select Arabic for our experiments: it is an important language with very limited resources.

This paper is organized as follows. Section II introduces some related works on EA. Section III describes the methodology we follow. The experiments and results are presented and discussed in Section IV. Finally, Section V concludes our work.

2. RELATED WORKS

In this section, we survey the literature focusing on three issues: automatic ways used for emotion labeling, EA of Arabic text and efforts on the use of ensemble classifiers.

2.1 Automatic Techniques for Emotions Labeling

---

[1] http://goo.gl/l3V9NZ

Many emotion classification methods need labeling the training data. Manual labeling is considered to be time-consuming, labor intensive and sometimes prone to errors. So, some studies employed Twitter to create a dataset with emotion labels using automatic methods such hashtags and emoticons found in tweets.

Purver and Battersby [4] trained their classifiers using automatically labeled data collected from Twitter. They used Ekman's six basic emotions and the collected tweets contained hashtags or emoticons corresponding to one of the six emotion categories. They found that the classifiers performed well on some emotion categories (anger, joy and sadness) and poor on other emotion categories (disgust, fear and surprise). Another study by Suttles and Ide [5] used Plutchik's eight basic emotions to collect emotional tweets. They labeled tweets automatically not only using hashtags and emoticons, but also using emojis. They showed that emojis may have more emotion indications than emoticons. Hasan et al. [6] used an automatic method to collect emotional tweets with no manual efforts in which 28 affect words from Circumplex model [7] are identified as seed keywords. Then, the keywords are extended using synonyms from WordNet lexicon and used as hashtags to collect emotional tweets. In addition, emoticons are used as features when training the machine learning classifiers.

While many studies employed emoticons for automatic labeling in English language, some studies used emoticons to label their data in other languages. Zhao et al. [8] focused on a Twitter-like social network in China called Weibo. They used emoticons to automatically collect Chinese tweets on four emotion categories (anger, disgust, joy and sadness). Then, the labeled data is used to train a Naive Bayes classifier.

Recently, Fernández-Gavilanes et al. [46] proposed an interesting approach to create emoji lexicons in an automatic unsupervised way. Following the same approach as ours, they collected emojis and their definitions from emojipedia and considered the sentiment distribution of the informal texts accompanying these emojis to construct sentiment lexicons. They showed the effectiveness of their lexicons on English and Spanish datasets.

2.2 Emotions Analysis in the Arabic Language

Most of the existing researches on EA and SA focused on the English language while few studies handled the problems of EA and SA for the Arabic language. Many studies were conducted on SA of Arabic posts over the past few years. We are not going to discuss them here. Interested readers are referred to recent surveys [10-12] for a comprehensive coverage of these studies. This subsection focuses mainly on EA studies conducted for the Arabic language.

One of the earliest works on EA of Arabic text was conducted by El Gohary et al. [13]. The authors constructed a model to detect emotions in Arabic children stories using a lexicon-based approach. The model they used is based on a moderate size lexicon to identify the emotional expressions at different levels (word, sentence and document). El Gohary et al. dataset consists of 100 documents (2,514 sentences) where 65 documents were used for training and the rest of the documents for testing. The training dataset is annotated manually using the six basic emotions of Ekman (joy, sadness, anger, fear, disgust and surprise), in addition to two more categories: neutral category (carried no emotion) and mixed (carried more than one emotion). They started their approach with preprocessing the text to remove stop words and perform stemming. After performing the preprocessing steps, the Vector Space Model (VSM) is built to measure the similarities between the sentences and the constructed six emotion lexicons using Cosine measure.

Omneya and Sturm [14] showed that emotions can be detected automatically from Arabic tweets after applying Arabic preprocessing. They started their work collecting 1,776 tweets related to the Egyptian revolution in 2011 and manually annotated each tweet using Ekman's six basic emotions in addition to a lexicon of words for each emotion extracted from the dataset. The annotation process is performed by labeling each tweet with corresponding emotion category using an average of 15 human annotators. They excluded any tweet with agreement less than 50% between annotators. So, only 1,605 tweets are left. The authors performed basic preprocessing steps (to remove non-Arabic letters, punctuations and multiple spaces) in addition to the removal of stop words and stemming. They use the unigram model to represent their dataset and use the WEKA software as a suite of machine learning algorithms to find the sets of words that are highly correlated with each emotion from the first 1,012 annotated tweets. These lists represent a seed for emotional lexicons that were later expanded manually. Finally, the authors compared the effectiveness of two common classifiers, which are Support Vector Machine (SVM) and Naïve Bayes (NB).

Abd Al-Aziz et al. [15] proposed a method to detect emotions using a combination of lexicon-based approach and Multi-Criteria Decision Making (MCDM) approach. Their dataset consisted of 1,552 tweets collected from two different sources and the five lexicons of emotion words are manually constructed for each of the five emotion classes (happiness, sadness, anger, fear and disgust). Then, the five emotion lexicons are evaluated by two judges based on some criteria and Cohen's Kappa coefficient is performed to measure the agreements between the two judges. After that, based on the emotion lexicons, the emotion scoring algorithm is used to represent each tweet as a vector consisting of five emotion scores. Finally, Co-Plot, which is a special form of MCDM, is used to classify tweets to their emotion state using 2-D graphical representation. The importance of this approach is in its ability to handle tweets with mixed emotions.

Al-A'abed and Al-Ayyoub [16] developed a lexicon-based approach to detect emotions from Arabic text. They make use of an existing emotion lexicon called EmoLex [17] which considers Plutchik's eight emotions. Primarily, EmoLexis was created for English and consisted of 14,182 terms. It was later translated to 20 different languages including Arabic. After removing terms conveying no emotions and duplicates caused by translation, the lexicon is reduced to 4,279 Arabic terms which is also used in [18, 43]. Finally, they evaluate their approach using 39 text excerpts collected from different online resources and achieved an accuracy of 89.7%.

In perhaps the closest work to ours, Refaee and Rieser [9, 33-37] followed a DS learning approach and used emoticons to automatically label Arabic tweets collected from Twitter. While being very interesting and pioneering, the work of Refaee and Rieser differ from ours in many aspects. First, they focused on SA whereas we focus on EA, which is more challenging as discussed earlier. The complexity of EA might be why Refaee and Rieser state that "the vast majority of previous work has used DS only with binary sentiment classification positive vs. negative" [37]. Another difference is the way emoticons are used. Refaee and Rieser used a very limited set of emoticons (six positive and seven negative) whereas we use a more extensive set. Moreover, they did not really discuss in details how the automatic annotation was done. E.g., they did not show how they handle tweets with mixed/conflicting emoticons. One more difference is related to the ratio between the automatically annotated training dataset and the manually annotated one is much larger in their work compared to ours. In their work, this ratio exceeds 60, whereas, in our work, it is only around 10. The final issue is related reproducibility. Refaee and Rieser employed complex features (including syntactic, semantic, morphological, language-style and Twitter-specific features) extracted using tools (such as MADAMIRA [44]) that are no longer publicly available.[2] Moreover, the links they provided to access their resources are dead, whereas our work employs simple features and we share our resources on publicly accessible repository.[3]

2.3 Combining Classifiers

The combination of multiple classifiers to produce a single classifier has been an active area of study over the previous two decades [19-21]. Many studies show that combining classifiers can improve the results of classification. The advantage of combining classifiers is to exploit the strengths of individual classifiers and to evade their weakness, resulting in the enhancement of classification accuracy [22].

Y. Bi [22] investigates the combination of four different machine learning algorithms for text categorization using Dempster combination rule. The four learning algorithms include Support Vector Machine (SVM), kNN (Nearest Neighbor), kNN model-based approach (kNNM), and Rocchio. First, they trained each classifier individually using two datasets: 20-newsgroup and Reuters-21578 benchmark. Second, the same datasets are used to evaluate the combinations of multiple classifiers and compare the results with individual classifiers results. Finally, their experiments showed that the result of best-combined classifiers improves the performance of the best individual classifier (which is SVM).

Xia et al. [23] examined the performance of numerous linguistic features of text documents, including part of speech and term frequency - inverse document frequency (TF-IDF), in combination with ensemble methods for sentiment classification. In their scheme, several classification algorithms such as support vector machines (SVM), Naïve Bayes (NB) and maximum entropy (ME) were combined by fixed combining rules, meta-classifier, and weighted combination rules.

Recently, Da Silva et al. [24] examined the performance of two feature schemes, which are feature hashing and bag-of-words, in sentiment analysis on data from Twitter. In their experiments, different classification schemes were conducted with the combination of bag-of-words, lexicons, emoticons and feature hashing. They used four machine learning classifiers; which are support vector machines (SVM),

---

[2] MADAMIRA's website (https://camel.abudhabi.nyu.edu/madamira/) does have a download link. Unfortunately, clicking on it re-directs the user to another page stating that the "Licensing Status" is "Available for licensing and sponsored research support".

[3] https://github.com/malayyoub/emojis-to-train-emotion-classifiers

TABLE II. SAMPLE OF LABELING TWEET BASED ON EXISTING EMOJIS.

| # | Tweet | Category |
|---|---|---|
| 1 | RT @777Ropi # قلة الادب والوقاحه وعدم الاحترام خصوصا مع كبار السن ☹️ شي_ينرفزك_ #It_makes_me_angry lack of manners, rudeness and disrespect, especially with the elderly. | (☹️ = -5) anger |
| 2 | الله يرحمه ويغفر له ويسكنه فسيح جناته ياررب💔. #استشهاد_الملازم_فيصل_الطوب May Allah bless his soul, forgive him and put him in paradise. #Martyrdom_lieutenant_Faisal_Altoob | (💔 = -3) sadness |
| 3 | ❤️❤️😊زد_رصيدك59 مايعجبني من البرنامج بكبره الا فقرة الشيخ مشاري الله يجزاه الجنة # What I like about the whole program is only a scene of Sheikh Mishari, may Allah put him in paradise. | (😊 = -4) (❤️❤️=3*2) joy |

Naïve Bayes (NB), random forest (RF), and logistic regression (LR); were combined to construct a classifier ensemble. The ensemble results indicate that combining classifiers can improve the classification performance.

3. METHODOLOGY

This section explains in details the steps followed to accomplish this study. First, the data collection and labeling approaches are presented. Next, preprocessing and feature extraction are presented. The Bag-Of-Words (BOW) model includes a tokenization of text to extract terms (words) and construct feature vectors that based on the frequency of each extracted term (word). Once the features are extracted, the manually labeled data is split into a training set and a testing set. Each approach is trained using classification algorithm that builds a model of it. Finally, the same testing set is used to evaluate the models of the two labeling approaches.

3.1 COLLECTING DATA

We collect our data from Twitter using trending hashtags. It consists of 134,194 Arabic tweets collected from August 2015 to February 2016. Using emoticons can carry strong sentiment [25] and some users express their opinions and moods using emoticons when posting tweets. A large number of Twitter users use small digital images called emojis which are a new generation of emoticons represented as Unicode symbols. Since we only focus on four categories of emotions: anger, disgust, joy and sadness, we select emojis that are related to these four emotion categories. There are 845 emojis in Twitter, but we only consider the top used emojis. First, we scan through the dataset in order to estimate the top frequently used emojis. Afterward, these emojis are categorized into one of the four categories of emotions adopted in our study. Table I exhibits a sample of emojis distributed among the four categories of emotions.

3.2 Data Labeling Approaches

In this study, we present two approaches of labeling the data: automatic labeling and manual labeling.

3.2.1 Automatic Labeling

After assigning the emojis into their categories, we label the tweet according to the emojis it contains. The AFINN lexicon [26] is used in the process of determining the main emotion in each tweet (in case the tweet contains emojis representing different emotions. The AFINN lexicon is a list of English words that are manually labeled and assigned a weight (score) that ranges between -5 and +5, where the negative scores represent negative sentiments and the positive scores represents positive sentiments. In addition, there is a list of emojis with their weight (scores) listed in the AFINN lexicon.

TABLE III. THE NUMBER OF TWEETS IN EACH CATEGORY OF AUTOMATICALLY LABELED DATA.

| Category | Total number of tweets |
|---|---|
| **Joy** | 10467 |
| **Sadness** | 7878 |
| **Anger** | 2874 |
| **Disgust** | 1533 |
| **Total** | 22752 |

TABLE IV. THE NUMBER OF TWEETS IN EACH CATEGORY OF MANUALLY LABELED DATA WITH TRAIN/TEST PERCENTAGES.

| Category | Number of tweets | Training (80%) | Testing (20%) |
|---|---|---|---|
| Joy | 630 | 504 | 126 |
| Sadness | 415 | 332 | 83 |
| Anger | 620 | 496 | 124 |
| Disgust | 360 | 288 | 72 |
| Total | 2025 | 1620 | 405 |

We use the emojis in AFINN lexicon in addition to other emojis that are not listed in AFINN lexicon. To assign scores to these additional emojis, we need to do further research. Specifally, we investigat each emoji and extract its name name, description and usage as presented in emojipedia[4]. We then search for the emoji's equivalent name from the AFINN lexicon and assign it its score to the emoji. For example, the broken heart emoji (💔) does not exist in AFINN lexicon but it is used a lot to express sadness emotion. So, we assign to it the score of the word "heartbroken" which exists in the AFINN lexicon.

Our automatic annotation approach is as follows. For each tweet, we go over each emoji it contains and look up the emotion it conveys as well as its weight (score). We sum up the weights (scores) of the emojis pertaining to each emotion and simply choose the emotion with the highest weight (score). Ties are very rare and are handled randomly.

Table 2 shows some examples of the way we label the tweets based on the emojis they contain. The first tweet is labeled with Anger due to the presence of an anger emoji with a score of -5 as indicated in the lexicon. The second tweet is labeled with Sadness due to the presence of a sadness emoji with a score of -3. The third tweet is considered as a Joy tweet due to the presence of two Joy emojis (each with a score of 3) and one Sadness emoji (with a score of -4), which means that the total score for Joy is higher than Sadness.

We have collected 134,194 tweets using a crawler built using C# programming language. Then, we extract 22,752 tweets that contain emojis and automatically determine the emotion of each tweet using the technique described in the previous paragraph. These tweets are distributed among the four emotion categories under consideration as shown in Table 3.

*3.2.2 MANUAL LABELING*
On the other hand, we select 2025 tweets from our collected data that are free from any emojis and manually label them into the four emotion categories mentioned before. Afterward, these tweets are divided into 1620 tweets as training data that constitute (80%) of the 2025 tweets, and 405 tweets as test data that constitute (20%) of the tweets under consideration. Table 4 shows the number of tweets in each category of emotion with train/test percentages.

3.3 Preprocessing
Most of the previous studies on EA were based on one of the two approaches: the unsupervised approach (lexicon-based) and the supervised approach (machine learning-based). Therefore, no matter which approach is applied, the prerequisite for text pre-processing tasks is crucial and inevitable to make the inputted text readable for both approaches. To improve the efficiency of our work, several preprocessing steps are employed.

3.3.1 Removing Hashtags at The End
In Twitter, hash-tags can be placed in the beginning, middle, or end of a tweet. As part of pre-processing, the tags in the beginning or in the middle of the tweet are left, since they are part of the sentence. For example, the tweet "حادثه #الألعاب_الناريه حادث مأساوي ينم عن استهتار وفساد", which means "The accident of fireworks is tragic and tell about disrespect and corruption", we can notice that the hashtag "#الألعاب_الناريه" is actually part of the tweet and if we remove it, the tweet would be meaningless. While hash-tags at the end of tweets are removed to reduce the features dimensionality [6].

---
[4] http://www.emojipedia.org

3.3.2 Repeated Letters Elimination
Words with repeated letters such as "مبروووووك", which means "congratulation", are common in Twitter messages. Any letter occurring more than two times consecutively is replaced with one occurrence. For instance, the word "مبروووووك" would be changed into "مبروك".

3.3.3 Normalization
The Arabic language is known as a morphological language. Letters in the Arabic language can be written in different shapes and forms based on their location in the word. For example, the letter "Alif" has four forms ('آ', 'إ', 'أ', 'ا'). Hence, to avoid the performance degradation and improve the classification, normalizing some letters such as 'آ', 'إ', 'أ' to 'ا' and 'ة' to 'ه' become vital.

3.3.4 Stemming
Stemming, which is also called Lemmatization, is the process of reducing inflected words into their main root or stem. For instance, in the Arabic language the three words "يكتب", "الكاتب", "الكتابة" which mean "the writing", "the writer", and "write" are all reduced to the stem "كتب" which means "wrote". Word stemming is an important pre-processing step in Arabic text classification [27], which might be beneficial in our study since EA is part of text classification.
Currently, there are two morphological analysis methods for the Arabic language. The first method is the root-based stemming, which removes all affixes and returns each inputted Arabic word to its root form [28]. The second method is the light stemming, which remove only the common affixes (prefixes and suffixes) without modifying the origin (root) of a word [27]. In our study, we applied only the light stemming method.

3.3.5 Stop-Words Removal
Every language has its own stop-words list which should be filtered out from a text during the pre-processing stage. Generally, stop-words list contains pronouns, prepositions, articles, etc. [29]. In Arabic language, stop-words list could compose of pronouns ("الذي", "هم", "هو", etc. which mean "he", "they", "which"), prepositions ("على", "في", "من", etc. which mean "from", "in", "on"), punctuations ('!' ',' '?' ';' etc.), in addition to other words like the names of days and months [27]. For the EA problem, stop-words have no valuable information. That is, a preposition, for example, does not convey any emotion about anything. Therefore, it is better to remove stop-words before the classification. This step reduces the features dimensionality and decreases storage.

3.4 FEATURE EXTRACTION

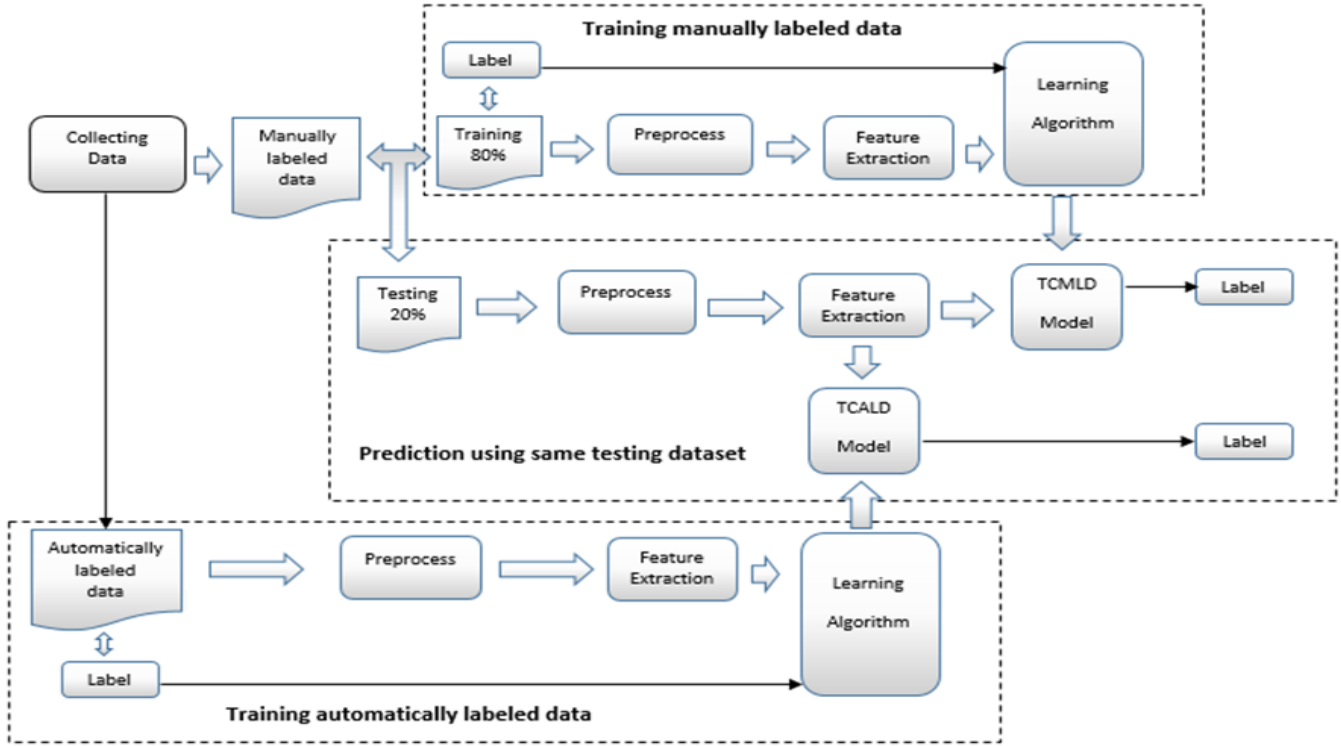

*Fig. 1. Methodology*

We used the BOW approach as feature extraction. BOW relies on the occurrences of terms within a document. To provide more meaningful features, the term frequency-inverse document frequency (TF-IDF) weighting technique is applied which yields higher weights to terms appearing frequently in the given tweet and rarely in other tweets.

*3.5 SYSTEM ARCHITECTURE*

After collecting our dataset as mentioned earlier, the manually labeled data is split into training dataset (80%) and testing dataset (20%). First, the training dataset is preprocessed and the feature extraction is performed. Then, the machine learning classifier is learned and produced a trained model called Trained Classifier for Manually Labeled Data (TCMLD). After producing the model, the testing dataset would also be preprocessed and the feature would be extracted. Finally, the testing dataset will be used to evaluate TCMLD model.

On the other hand, the automatically labeled data is also preprocessed and the feature extraction is performed. Then, the machine learning classifier is learned using the automatically labeled data and produced a trained model called Trained Classifier for Automatically Labeled Data (TCALD). Finally, the testing dataset used to evaluate TCMLD model will also be used to evaluate the TCALD model. Figure 1 shows in details the classification of manually and automatically labeled data using the same testing dataset.

*3.6 CLASSIFICATION*

Generally, Naive Bayes (NB), Support Vector Machine (SVM) and Random Forest (RF) classifiers are known to perform very well for the general text classification problem [30]. Since we are dealing with a multi-class problem in this study, we found out that Multinomial Naive Bayes (MNB) outperformed NB with multi-class text classification. Therefore, for the feature extraction approach under consideration, we experimented with the following three classifiers:
- Support Vector Machine (SVM).
- Multinomial Naive Bayes (MNB).
- Random Forest (RF).

Since text classification depends significantly on training and testing data, so to be fair, each experiment is conducted five times, wherein each time we change the data of training and testing. This is known as five folds cross validation.

3.7 Combining Classifiers

Combining classifiers, which is also called ensemble of learners, considered to be a significant subfield of machine learning. It aims to build classification models to improve the predictive performance by combining the decisions of multiple learning classifiers. Different methods of combining classifiers are proposed but we will focus only on fixed rules method since we use it in this study.

3.7.1 Fixed Combining Rules

Fixed combining rules are considered to be the simplest combination approach and it is the most widely used in the multiple classifier system [31]. The fixed combining rules produce the classification output by combining the results of multiple classifiers. In this approach, all classifiers in the model are trained separately and each classifier provides its decision, then the results of the classifiers are combined to give the final decision according to the rule used in combination. Many rules can be used to combine the results, the four rules; average rule, product rule, maximum rule and minimum rule [32]; are defined as follows.

- The average rule: The average rule or the average of probabilities can be defined as finding the maximum value of the average of all probabilities produced for each class.
- The product rule: The product rule or the product of probabilities multiplies the probabilities provided by each individual classifier and assign the input text to the class with the maximum value.
- The maximum rule: The maximum rule or the maximum of probabilities finds the maximum probability value for each class between the individual classifiers and assign the input text to the class with the maximum value.
- The minimum rule: The minimum rule or the minimum of probabilities finds the minimum probability value for each class between the individual classifiers and assign the input text to the class with the maximum value.

4. EXPERIMENTS AND RESULTS

In our experiments, we aim at evaluating the effectiveness of the automatic labeling approach of the training data compared with the manual approach. We study the effect of the preprocessing steps and the use of single-emoji data for the automatic approach. Moreover, we experiment with individual classifiers as well as ensemble classifiers. For evaluation, we report the weighted averages of precision, recall and f-measure.

4.1 Using Individual Classifiers

In this section, we discuss the experiments we conduct using individual classifiers. Figures 3-5 show the classification results of training SVM, MNB and RF, respectively, on the manually and automatically labeled data. These results show clearly that the classifiers under consideration exhibit better performance on trained on the automatically labeled data compared with the manually labeled data.

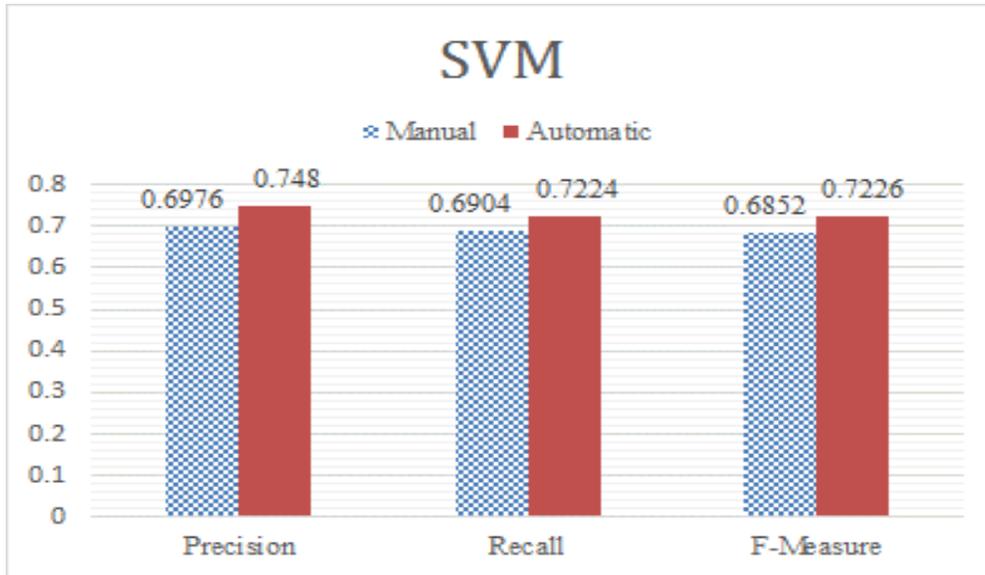

**Figure 3:** Precision, Recall and F-measure for training SVM using automatically labeled data and manually labeled data.

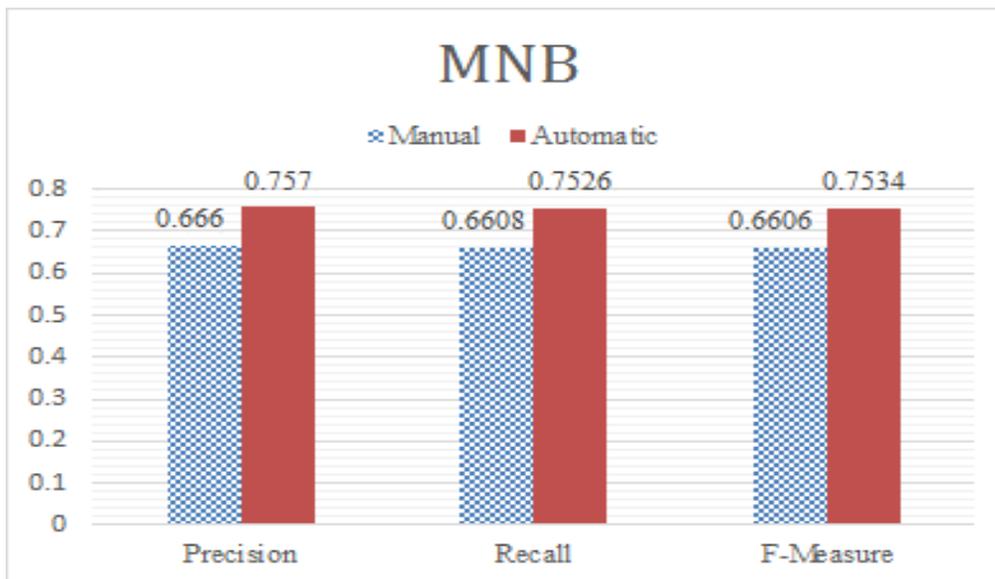

**Figure 4:** Precision, Recall and F-measure for training MNB using automatically labeled data and manually labeled data.

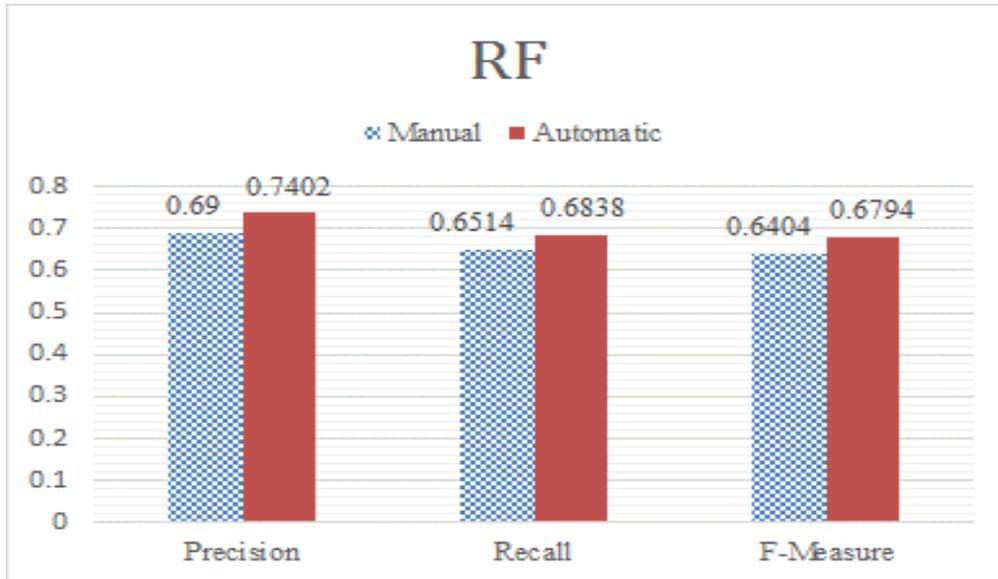

**Figure 5:** Precision, recall and f-measure for training RF using automatically labeled data and manually labeled data.

4.2 Measuring the Effect of Preprocessing

Earlier work, such as that of Refaee and Rieser [9, 33-37], reported that preprocessing steps may have a significant effect on the results. Thus, we discuss in this section the experiments we conduct to measure the effect of preprocessing steps such as stemming, normalization, etc. Figures 6-8 show the classification results of training the manually and automatically labeled data using SVM, MNB and RF, respectively, without performing the preprocessing steps. As observed in these figures, a significant drop in accuracy is suffered by all classifiers when fed un-preprocessed data. Moreover, the results show that effect or not performing the preprocessing steps varies across the combinations of the two approaches and three classifiers under consideration. For example, MNB is the most robust classifiers since its performance is least affected in this setting.

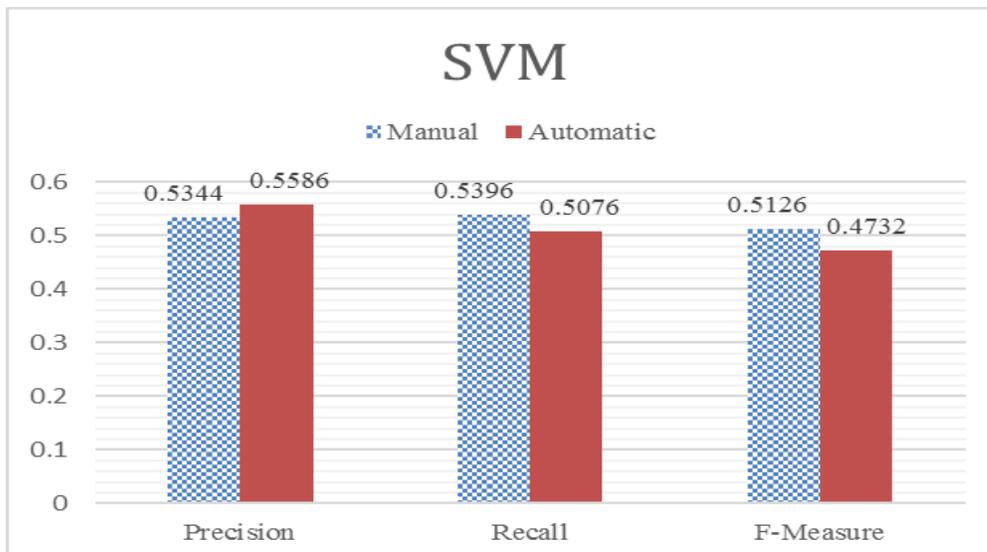

**Figure 6:** Precision, Recall and F-measure for training SVM using automatically labeled un-preprocessed data and manually labeled un-preprocessed data.

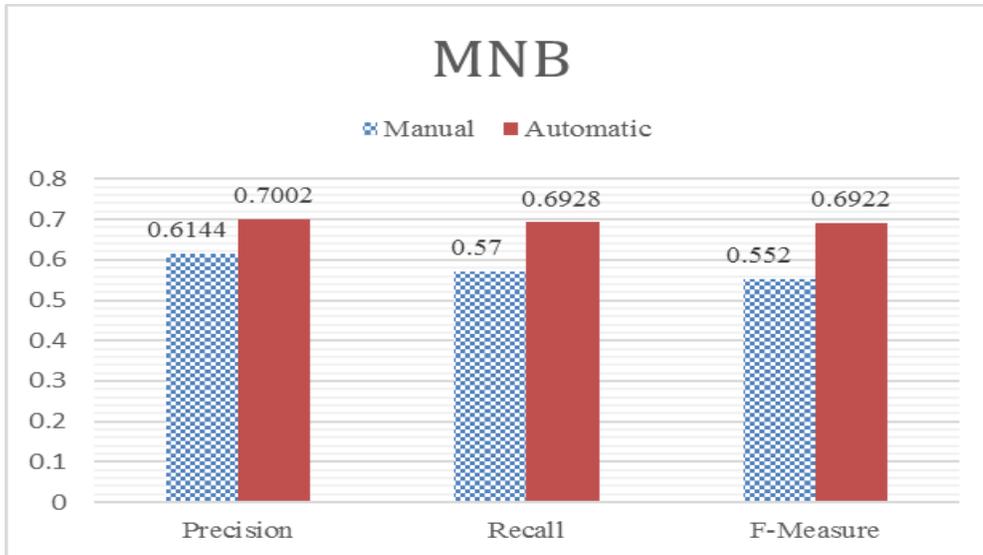

**Figure 7:** Precision, Recall and F-measure for training MNB using automatically labeled un-preprocessed data and manually labeled un-preprocessed data.

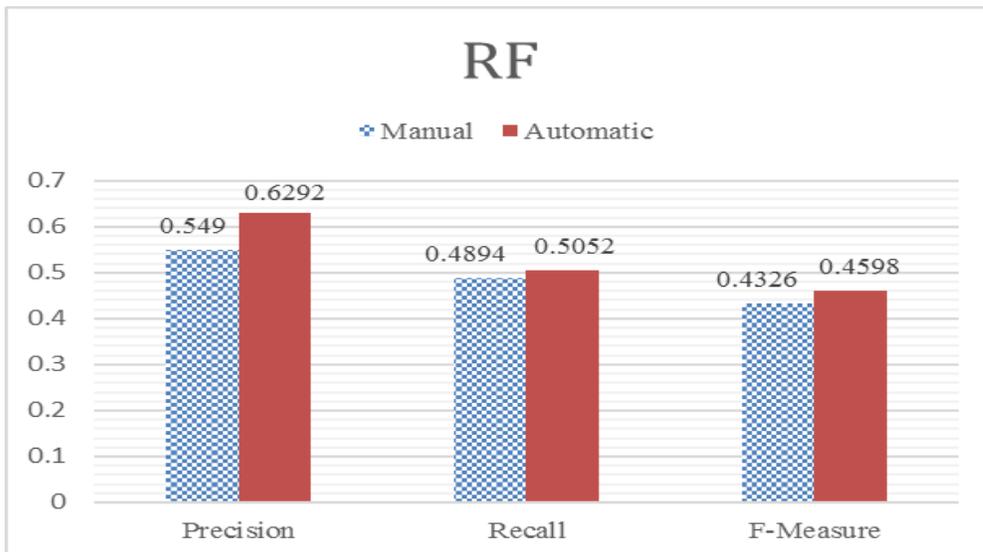

**Figure 8:** Precision, Recall and F-measure for training RF using automatically labeled un-preprocessed data and manually labeled un-preprocessed data.

4.3 Measuring the Effect of Using Single-Emoji Tweets for TALD

Another issue discussed in the DS literature is the level of noise in the automatically generated/labeled training data. In our case, one would assume that a tweet with a single emoji would convey simpler and purer emotions, which makes it less "noisy" for the purposes of DS learning. To test this claim, we filter out the tweets with multiple emojis, which represent around 42.5% of the automatically labeled tweets, and study the TALD approach with this data. Figures 9-11 show the effect of using single-emoji tweets for training SVM, MNB and RF, respectively, using the automatic approach. The results show that limiting the training data to only single-emoji tweets did not affect the results of any classifier in any statistically significant way.

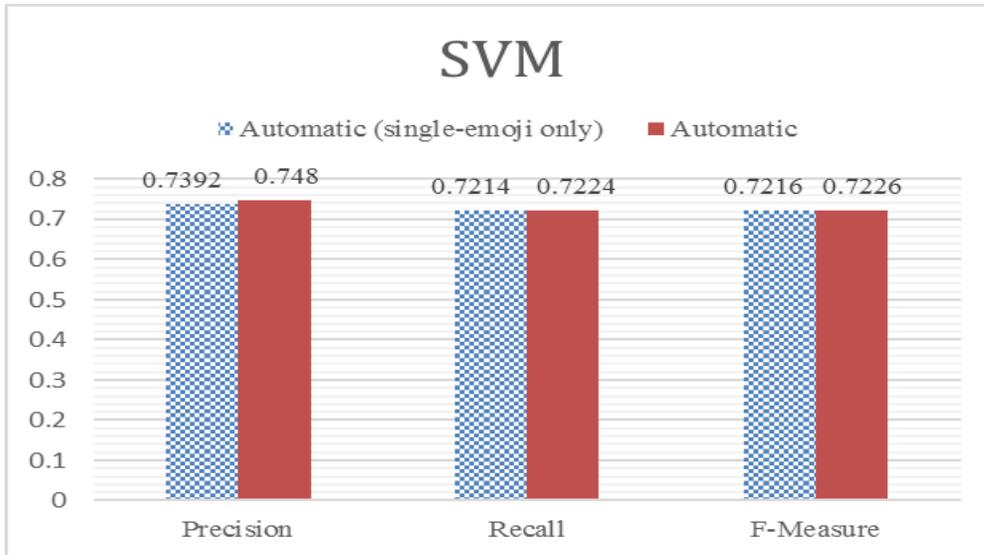

**Figure 9:** The effect of using single-emoji tweets for training SVM using the automatic approach.

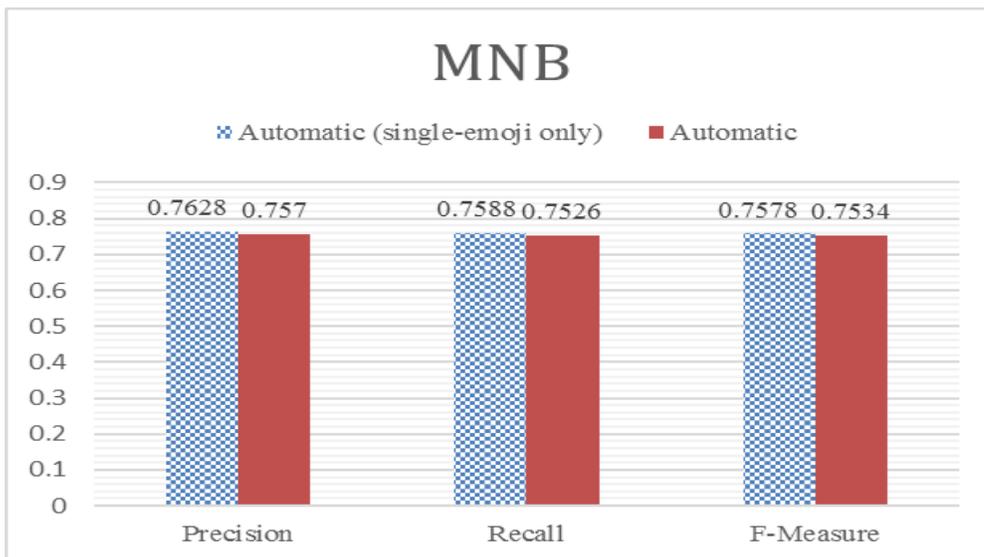

**Figure 10:** The effect of using single-emoji tweets for training MNB using the automatic approach.

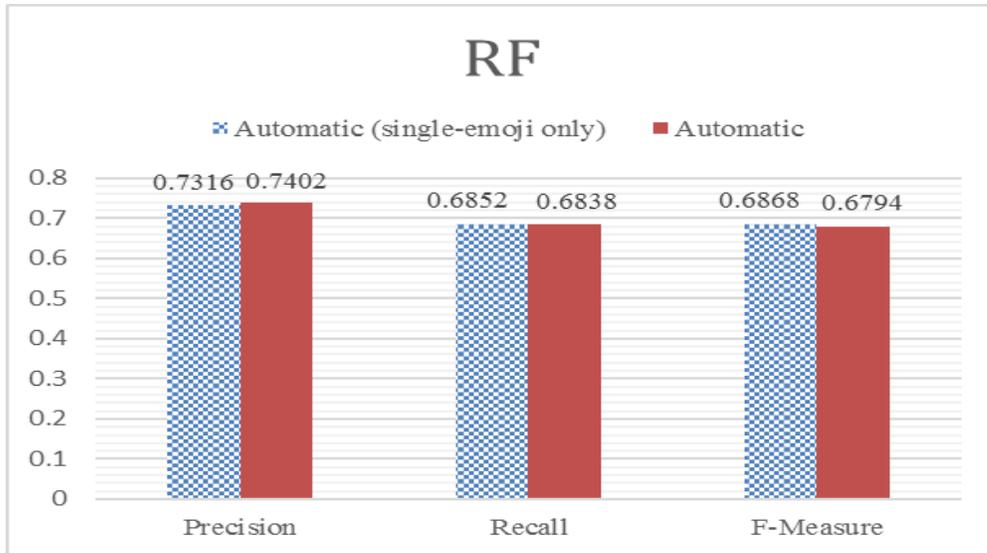

**Figure 11:** The effect of using single-emoji tweets for training RF using the automatic approach.

4.2 Experiments on Combining Classifiers

For combining classifiers, we use four different fixed combination rules:

1-      Average rule (average of probabilities).

2-      Product rule (product of probabilities).

3-      Maximum rule (maximum of probabilities).

4-      Minimum rule (minimum of probabilities).

We apply these four fixed rules using three classification algorithms (SVM, MNB and RF) for results of both TALD and TMLD aiming to improve the classification accuracy. In the fixed rule combining method, each classifier produces its output (probability distribution for each class) separately, then the final output of combination is computed based on the rule used. Figure 6 provides an example of combining the outputs of classifiers using the fixed combining rules mentioned above. Since we are dealing with four emotion classes, so for each tested tweet each classifier will produce the probability for each class and accordingly label the tweet based on the maximum class. The tweet in Figure 6 is originally labeled as class "anger", but only MNB classifier predicts the class correctly whereas SVM and RF predict the class incorrectly as "joy". Hence, we combine the outputs of the three classifiers using four fixed rules and we can see from Figure 6 that each rule gives the output as class "anger" which is the correct class for the tweet.

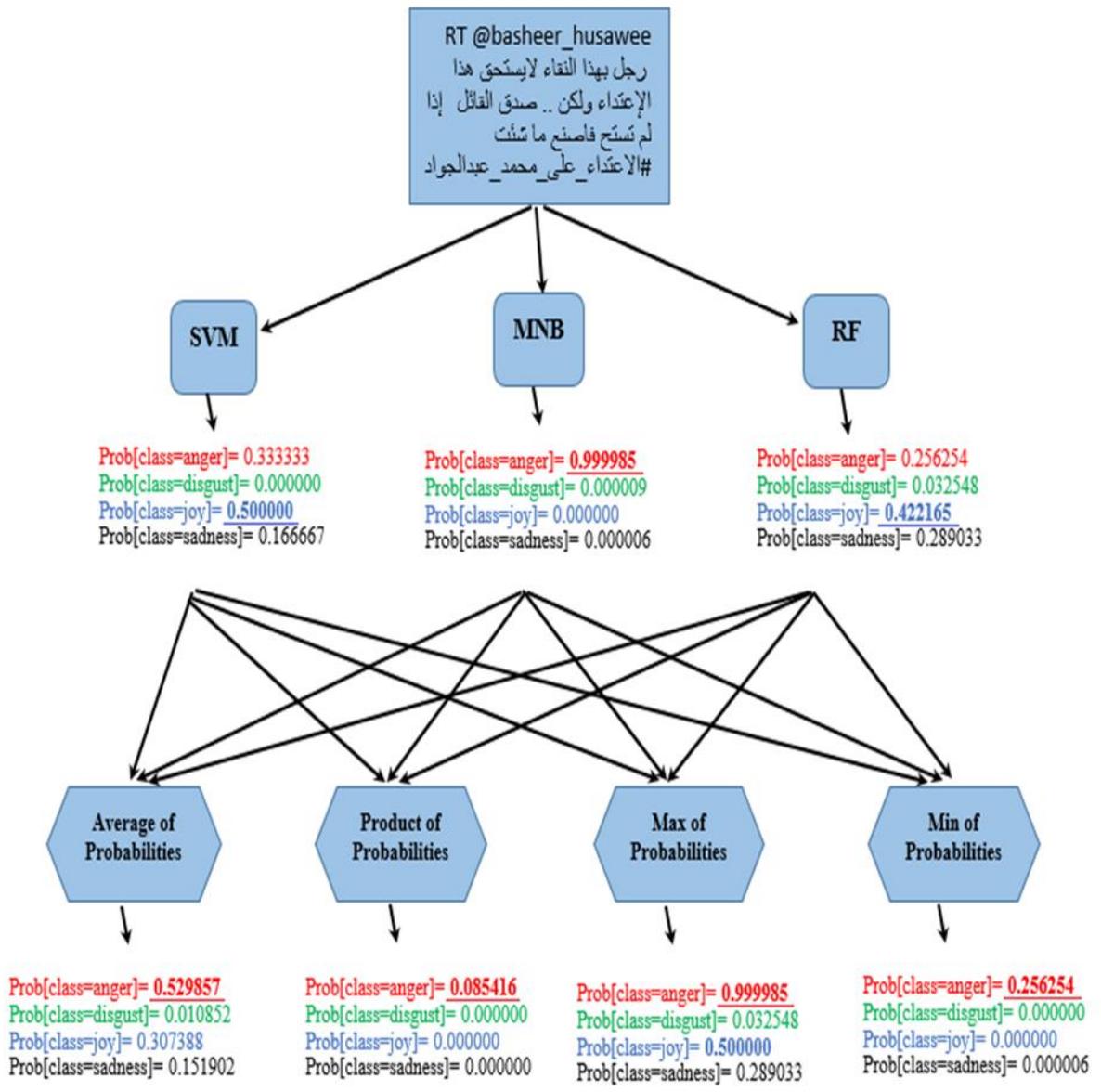

Figure 6: An example of combining SVM, MNB and RF classifiers using four fixed rules.

4.2.1 Results of Combining Classifiers of Automatic Labeling Approach

For evaluating the combining method, two experiments are conducted in this study and five folds for each experiment. The first experiment is conducted to combine the classification results of TALD, while, the second experiment is conducted to combine the classification results of TMLD. Both TALD and TMLD are combined using four combining rules. In this subsection, we present the results of combining classifiers of TALD and compare them with results of individual classifiers.

Table 5 shows the results of combining three classifiers (SVM, MNB and RF) using four combining rules, the results shows that applying the minimum rule achieved the highest accuracy (77.62964%).

Table 5: Accuracy of combining three classifiers using four combining rules.

|  | Combining SVM, MNB and RF | | | |
|---|---|---|---|---|
|  | **Average** | **Product** | **Maximum** | **Minimum** |
| 1st fold | 75.8 | 77 | 74.07 | 77.28 |
| 2nd fold | 75.8 | 75.31 | 74.07 | 75.8 |
| 3rd fold | 76.79 | 78.27 | 76.3 | 78.77 |
| 4th fold | 71.85 | 72.84 | 70.86 | 72.59 |
| 5th fold | 84.2 | 83.95 | 83.21 | 83.70 |
| Avg. | **76.89** | **77.481** | **75.70** | **77.63** |

4.2.1.1 Comparing Combined Method with Individual Classifiers

We compared the accuracy of combining classifiers that we get from combining rules with the accuracy of individual classifiers obtained previously in section 4.1 using TALD. Figure 7 shows the accuracy of combining three classifiers using four combining rules with individual classifiers.

From Figure 7 we notice that combining three classifiers (SVM, MNB and RF) give a high accuracy for all combining rules compared to the accuracy of individual classifiers used in TALD.

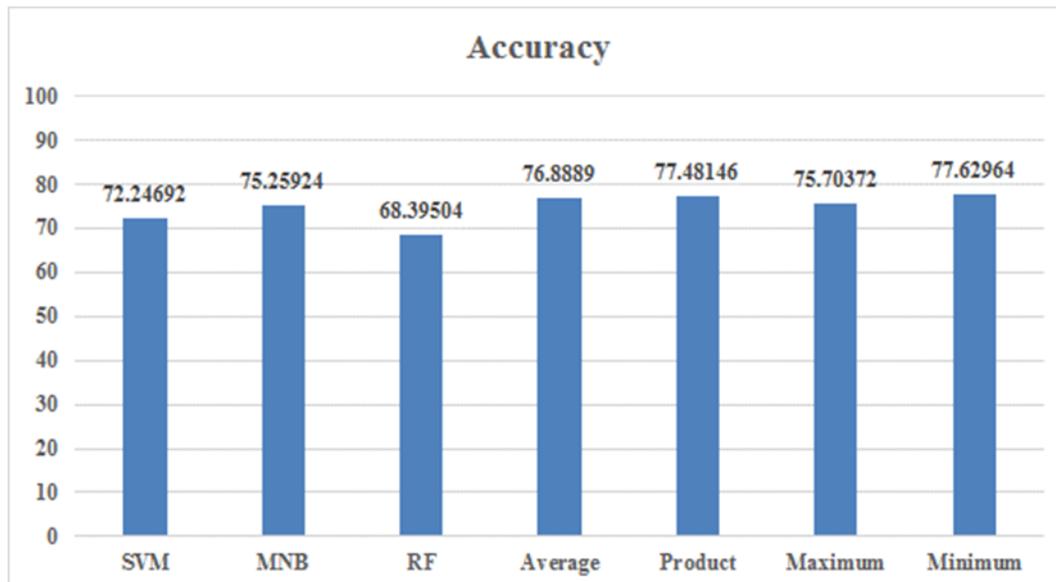

Figure 7: A comparison between three combined classifiers using four combining rules vs. individual classifiers of TALD.

4.2.2 Results of Combining Classifiers of Manual Labeling Approach

In this subsection, we present the results of combining classifiers of TMLD and compare them with results of individual classifiers.

Table 6 shows the results of combining three classifiers (SVM, MNB and RF) using four combining rules, the results show that applying the product rule achieved the highest accuracy (66.8642%).

Table 6: Accuracy of combining three classifiers using four combining rules.

|  | Combining SVM, MNB and RF | | | |
|---|---|---|---|---|
|  | Average | Product | Maximum | Minimum |
| 1st fold | 66.91 | 67.16 | 66.42 | 67.4 |
| 2nd fold | 66.67 | 67.9 | 66.42 | 67.9 |
| 3rd fold | 66.42 | 66.42 | 66.42 | 66.17 |
| 4th fold | 64.69 | 64.44 | 63.95 | 64.44 |
| 5th fold | 67.41 | 68.4 | 66.67 | 67.65 |
| Avg. | **66.42** | **66.86** | **65.98** | **66.72** |

4.2.2.1 Comparing Combined Method with Individual Classifiers

We compared the accuracy of combining classifiers that we get from combining rules with the accuracy of individual classifiers obtained previously in section 4.1 using TMLD. Figure 8 shows the accuracy of combining three classifiers using four combining rules with individual classifiers.

From Figure 8 we notice that the SVM classifier achieved individually the highest accuracy compared to the best of combining rule accuracy (the product rule).

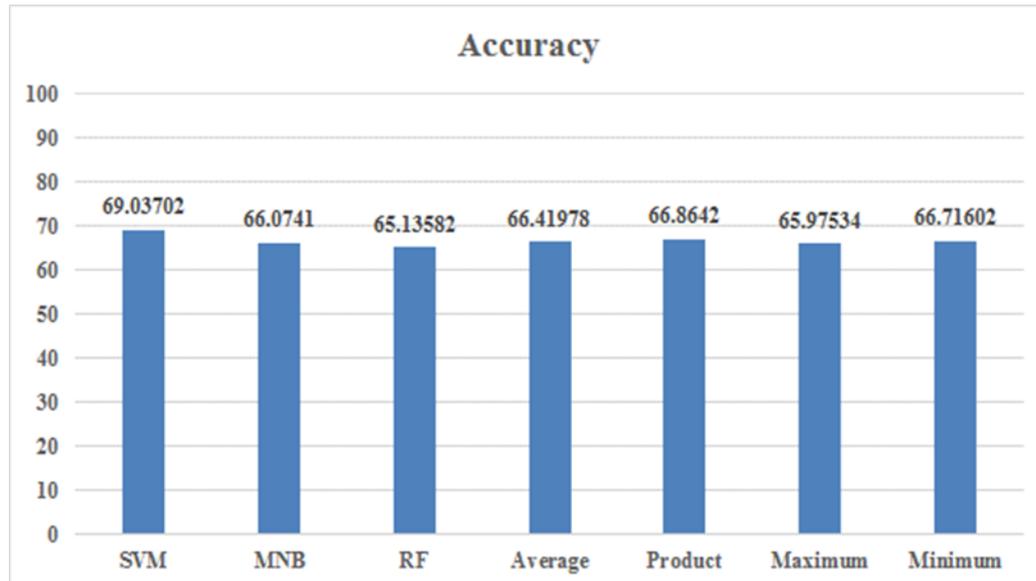

Figure 8: A comparison between three combined classifiers using four combining rules vs. individual classifiers of TMLD.

5. CONCLUSION

In this work, we study the problem of emotion analysis in Arabic social media. Twitter, which is one of the most popular micro-blogging services, is used to construct the dataset of Arabic tweets. We used emojis, which are a new generation of emoticons, to label the data automatically. A large number of Twitter users use emojis and recent studies have shown that the usage of emojis is growing and replacing emoticons on social media. In this study, we only focus on four categories of emotions: anger, disgust, joy and sadness, so we selected emojis that are related to these four emotion categories. There are 845 emojis in Twitter, but we only took into account the top used emojis. First, we investigated the collected dataset and count the frequency of each emoji to determine the top frequently used emojis. Out of 134,194 of Arabic tweets, we only extract 22,752 tweets that contain emojis and distributed them into their four emotion categories. In addition, we utilized AFINN lexicon scoring to guarantee assigning one category for each tweet in case different emojis exist in the tweet.

To test if labeling the data automatically using emojis is valid, we selected 2025 tweets from our collected data that are free from any emojis and manually labeled into the four emotion categories mentioned before. Labeling the data manually is considered time-consuming and labor intensive, so we only label a small amount of data as a prove of the concept. One fifth (20 %) of the manually labeled dataset is considered as a testing dataset to evaluate between both TMLD and TALD.

Three of known machine learning classifiers are considered in this study: Support Vector Machine (SVM), Multinomial Naive Bayes (MNB) and Random Forest (RF). First, the three classifiers are trained using the manually labeled data and evaluated using the testing dataset. Second, the same three classifiers are trained using the automatically labeled data and evaluated using the same testing dataset. Finally, the classification results show that TALD using SVM, MNB and RF is better than TMLD using the same classifiers.

Another goal of this study includes combining the outputs of classifiers in both TALD and TMLD in order to improve the accuracy of classification in each labeling approach. In this study, we combine the three classifiers (SVM, MNB and RF) and consider using four fixed combining rules which are average, product, maximum and minimum. First, the combining rules are applied to TALD, the combining results showed that all combining rules achieved high accuracies compared to the accuracies of individual classifiers. Second, the combining rules are applied to TMLD, the combining results showed that the SVM classifier achieved individually the highest accuracy compared to the best of combining rule accuracy.